%% file: iclr2024_conference.tex
\documentclass{article} 
\usepackage{iclr2024_conference,times}
\usepackage{algorithm}
\usepackage{algorithmic}
\usepackage{subcaption}

\input{math_commands.tex}

\usepackage{hyperref}
\usepackage{url}
\usepackage{graphicx}

\title{A tutorial note on collecting simulated data for vision-language-action models}

\author{Heran Wu, Zirun Zhou, Jingfeng Zhang* \\
School of Computer Science, 
The University of Auckland, New Zealand \\
*Contact Lecturer Jingfeng Zhang \texttt{<jingfeng.zhang9660@gmail.com>} for any questions
}

\iclrfinalcopy
\begin{document}
\maketitle

\begin{abstract}
Traditional robotic systems typically decompose intelligence into independent modules for computer vision, natural language processing, and motion control. Vision-Language-Action (VLA) models fundamentally transform this approach by employing a single neural network that can simultaneously process visual observations, understand human instructions, and directly output robot actions—all within a unified framework. However, these systems are highly dependent on high-quality training datasets that can capture the complex relationships between visual observations, language instructions, and robotic actions. This tutorial reviews three representative systems: the PyBullet simulation framework for flexible customized data generation, the LIBERO benchmark suite for standardized task definition and evaluation, and the RT-X dataset collection for large-scale multi-robot data acquisition. We demonstrated dataset generation approaches in PyBullet simulation and customized data collection within LIBERO, and provide an overview of the characteristics and roles of the RT-X dataset for large-scale multi-robot data acquisition. Code and data are available at https://github.com/trustmlyoungscientist/dataset\_for\_VLA.git.
\end{abstract}

\section*{Table of Contents}
\begin{itemize}
  \item \textbf{Vision-Language-Action Models: An Overview}
  \item \textbf{Data collection with PyBullet}
  \item \textbf{Data collection with Libero}
  \item \textbf{RT-X Dataset: Cross-Embodiment Learning for VLA}
  \item \textbf{Conclusion}
  \item \textbf{References}
\end{itemize}

\section{Vision-Language-Action Models: An Overview}

Vision-Language-Action (VLA) models have emerged as a transformative paradigm in robotics research, fundamentally changing how robots understand and execute tasks. Unlike traditional robotic systems that decompose intelligence into separate perception, planning, and control modules following the classical ``sense-plan-act'' paradigm, which decomposes robot intelligence into separate perception, planning, and control modules~\citep{brooks1986robust,murphy2000introduction},  VLA models learn unified end-to-end policies that simultaneously process visual observations, understand natural language instructions, and generate robotic actions.

Recent breakthroughs such as RT-1 (Robotics Transformer 1)~\citep{brohan2022rt}, RT-2~\citep{brohan2023rt2}, and OpenVLA~\citep{kim2024openvla} have demonstrated remarkable multi-task generalization capabilities, showcasing the potential of this unified approach. However, the success of these models critically depends on the quality and diversity of their training data—a challenge that distinguishes VLA research from traditional computer vision or natural language processing domains.

At its core, a VLA model learns a policy function that maps multimodal inputs to robotic actions:

\begin{equation}
\pi_\theta: (\boldsymbol{I}, \boldsymbol{L}) \rightarrow \mathcal{A},
\label{eq:policy}
\end{equation}

where $\boldsymbol{I}$ represents visual observations (RGB images), $\boldsymbol{L}$ represents language instructions, $\mathcal{A}$ represents robot actions, and $\theta$ denotes learnable parameters. This formulation enables robots to respond to natural commands like ``pick up the red block'' by processing camera imagery and text to generate precise joint positions, gripper commands, and motor signals.

Modern VLA systems typically consist of three interconnected components that process multimodal inputs in sequence. The \textbf{vision encoder} transforms raw camera observations into semantic representations:

\begin{equation}
\boldsymbol{h}_{\text{v}} = f_{\text{vision}}(\boldsymbol{I}) \in \mathbb{R}^{d_v},
\label{eq:vision_encoder}
\end{equation}

where $\boldsymbol{I}$ represents the visual observation.

The \textbf{language encoder} converts natural language instructions into dense semantic vectors:

\begin{equation}
\boldsymbol{h}_{\text{l}} = f_{\text{language}}(\boldsymbol{L}) \in \mathbb{R}^{d_l},
\label{eq:language_encoder}
\end{equation}

where $\boldsymbol{L}$ represents the language instruction.

Finally, the \textbf{policy network} fuses these multimodal representations to generate executable actions:

\begin{equation}
\boldsymbol{a}_t = \pi_\theta(\boldsymbol{h}_{\text{v}}, \boldsymbol{h}_{\text{l}}) \in \mathcal{A},
\label{eq:policy_network}
\end{equation}

where $\boldsymbol{a}_t$ represents the robot action at time step $t$, $\pi_\theta$ represents the parameterized policy function that takes visual and language features as input. $\boldsymbol{h}_{\text{v}}$ is the visual feature representation, $\boldsymbol{h}_{\text{l}}$ is the language feature representation, and $\mathcal{A}$ denotes the action space.

Training VLA models requires specialized approaches tailored to the multimodal nature of the problem. Behavior cloning~\citep{pomerleau1991efficient} frames policy learning as supervised learning from expert demonstrations, optimizing:
\begin{equation}
\mathcal{L}_{\text{BC}}(\theta) = \mathbb{E}_{(\boldsymbol{I},\boldsymbol{L},\mathcal{A}) \sim \mathcal{D}} \left[ \ell\left(\pi_\theta(\boldsymbol{I},\boldsymbol{L}), \mathcal{A}\right) \right],
\label{eq:behavior_cloning}
\end{equation}
where $\mathcal{L}_{\text{BC}}(\theta)$ is the behavior cloning loss function, $\theta$ denotes the policy parameters, $\mathcal{D}$ contains vision-language-action triplets, and $\ell$ represents an appropriate loss function.

Alternatively, \textbf{reinforcement learning} optimizes policies through environmental interaction:
\begin{equation}
J(\theta) = \mathbb{E}_{\tau \sim \pi_\theta} \left[ \sum_{t=0}^{T} \gamma^t r(s_t, a_t) \right],
\label{eq:reinforcement_learning}
\end{equation}

where $J(\theta)$ is the expected cumulative reward, $\theta$ denotes the policy parameters, $\tau$ represents a trajectory sampled by following policy $\pi_\theta$ to execute language instruction $\boldsymbol{L}$, $T$ is the trajectory length (number of time steps in $\tau$), $\gamma$ is the discount factor, $s_t$ is the state at time step $t$ (consisting of visual observations $\boldsymbol{I}_t$ and language instruction $\boldsymbol{L}$), $\boldsymbol{a}_t$ is the action at time step $t$ generated by $\pi_\theta(s_t, \boldsymbol{L})$, and $r(s_t, \boldsymbol{a}_t)$ is the reward function.

This tutorial addresses these challenges by examining three practical approaches: PyBullet for simulation-based data collection, RT-X for large-scale real robot data, and LIBERO for standardized benchmarks, providing concrete guidance for researchers tackling VLA dataset construction.

\section{Data Collection with PyBullet}

\subsection{PyBullet and Ravens Framework for VLA Dataset Construction}

PyBullet is a lightweight, Python-based physics simulation environment built on the Bullet physics engine~\citep{coumans2016pybullet}. Its ease of installation, straightforward API, and native support for URDF (Unified Robot Description Format) files make it an ideal choice for rapidly setting up manipulation scenarios in robotic learning. These features enabled us to quickly build customized manipulation tasks with fine-grained control over environment parameters, object properties, and data collection procedures.

For VLA dataset construction, we use PyBullet as the core physics simulation platform, combined with the Ravens framework~\citep{ravens2021framework} built on top of PyBullet as a structured task suite. Ravens provides predefined manipulation tasks, scripted oracle policies, and standardized evaluation metrics, while PyBullet handles the underlying physics simulation, camera rendering, and robot control mechanisms.

This combination brings multiple advantages to VLA research: PyBullet's flexible API allows precise control over simulation parameters and data collection procedures, while Ravens' structured task definitions ensure consistency and reproducibility across various manipulation scenarios. The scripted oracle policies provide deterministic, high-quality demonstration data, eliminating human operational randomness and serving as reliable benchmarks for model evaluation.

\subsection{Core Implementation Framework}

Our VLA dataset generation pipeline leverages PyBullet's simulation capabilities through the Ravens framework, focusing on systematic data collection across multiple task scenarios while maintaining controllability over simulation and data quality. Algorithm~\ref{alg:pybullet_pipeline} outlines the complete data generation process, which is implemented in the \texttt{ravens/demos.py} file in our repository.

\begin{figure}[htbp]
\centering
\includegraphics[width=\textwidth]{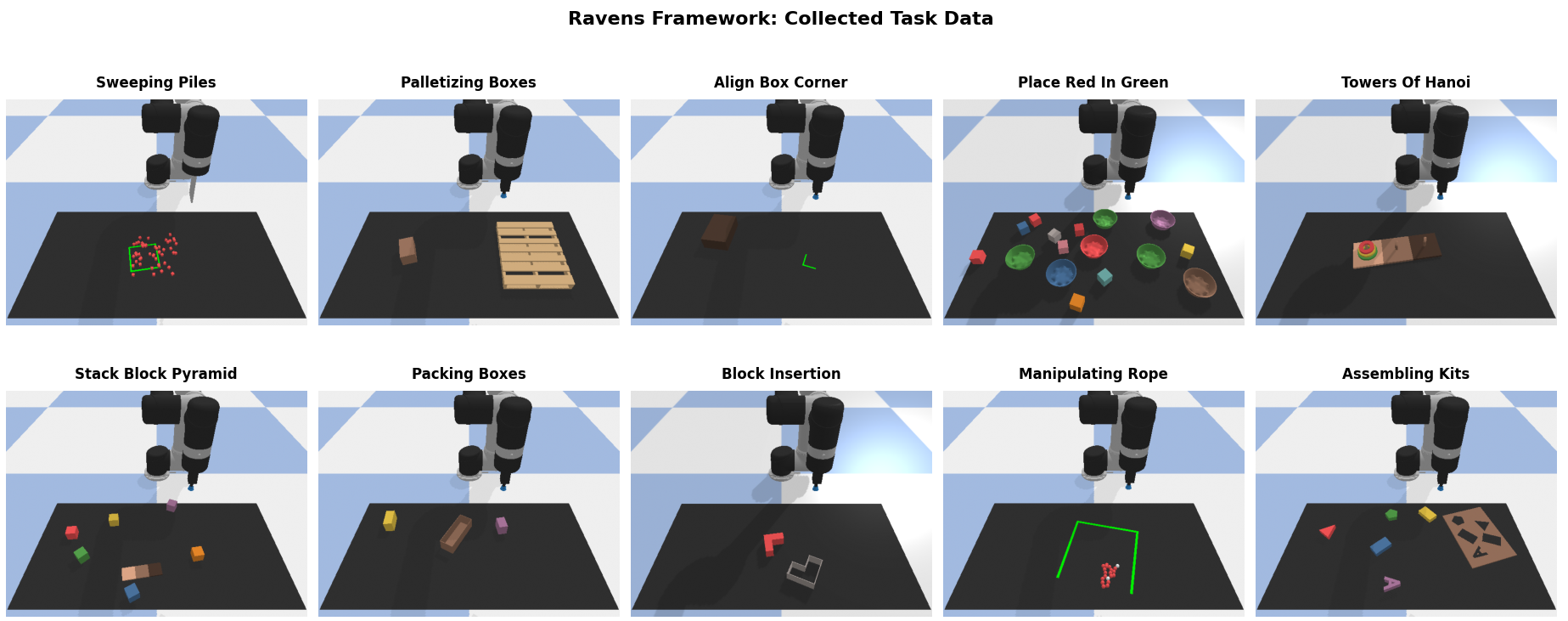}
\caption{Ravens Framework: Collected Task Data showing diverse manipulation scenarios in PyBullet simulation environment. 
The framework encompasses ten representative tasks including geometric alignment (\textit{Stack Block Pyramid}), perception-based sorting (\textit{Packing Boxes}), spatial reasoning (\textit{Align Box Corner}), multi-object coordination (\textit{Place Ball in Green}), sequential planning (\textit{Towers of Hanoi}), precision manipulation (\textit{Block Insertion}), assembly operations (\textit{Manipulating Rope}), and goal-directed placement tasks (\textit{Assembling Kits}). Each task provides systematic data collection for training VLA models across different manipulation skills and complexity levels.}
\label{fig:ravens_tasks}
\end{figure}

\begin{algorithm}
\caption{PyBullet VLA Data Generation Pipeline}
\label{alg:pybullet_pipeline}
\begin{algorithmic}[1]
\STATE env = ravens.Environment(assets\_root, disp=True) \COMMENT{Initialize Ravens environment}
\STATE tasks = [block-insertion, place-red-in-green, towers-of-hanoi]
\STATE p.connect(p.GUI) \COMMENT{Initialize PyBullet physics engine}
\FOR{task \textbf{in} tasks}
    \STATE env.set\_task(task) \COMMENT{Configure environment for current task}
    \FOR{episode \textbf{in} range(N)}
        \STATE obs = env.reset() \COMMENT{Reset environment and randomize scene}
        \STATE done = False
        \STATE step = 0
        \WHILE{\textbf{not} done}
            \STATE rgb\_img = p.getCameraImage(width, height, viewMatrix, projMatrix) \COMMENT{Capture RGB observation}
            \STATE action = oracle.act(obs) \COMMENT{Oracle policy generates expert action}
            \STATE obs, reward, done = env.step(action) \COMMENT{Execute action, get new state}
            \STATE save\_data(episode, step, rgb\_img, action, reward) \COMMENT{Store data triplet}
            \STATE step += 1
        \ENDWHILE
    \ENDFOR
\ENDFOR
\end{algorithmic}
\end{algorithm}

where \texttt{assets\_root} is the path to Ravens task assets, \texttt{N} is the number of episodes per task, \texttt{obs} represents the current observation including object poses, \texttt{oracle} is the scripted expert policy, and \texttt{save\_data()} is the function to store vision-language-action triplets.

\subsection{Task Selection and Data Characteristics}

We selected three representative manipulation tasks that demonstrate different types of robotic manipulation skills and task complexity levels:

Block-insertion tests precise geometric alignment and spatial reasoning capabilities. The task requires inserting an L-shaped block into a matching slot, demanding high-precision 6-degree-of-freedom (dof) pose estimation and fine control abilities.

Place-red-in-green evaluates color recognition and goal-directed placement capabilities. The task involves identifying objects through color attributes and executing conditional placement operations based on visual features.

Towers-of-hanoi assesses sequential reasoning and task planning capabilities. This classic puzzle requires understanding operational constraints, maintaining object ordering, and executing multi-step action sequences.

As shown in Figure~\ref{fig:ravens_tasks}, these tasks cover various primitive operations from pick-and-place and precise insertion to sequential object manipulation, embodying fundamental skills required for VLA models including visual perception, spatial understanding, and sequential planning.

We successfully collected demonstration data across three manipulation tasks. Three tasks achieved 95\% success rates, demonstrating the reliability of the PyBullet simulation environment and the stability of Ravens oracles. This provides a stable source of labeled data for subsequent model training and evaluation. Despite consistent success rates, tasks differ significantly in their underlying skill requirements. Block-insertion emphasizes geometric contact modeling and precise spatial alignment, place-red-in-green focuses on visual attribute understanding and object recognition, while towers-of-hanoi involves long-term planning and sequential constraint execution. This diversity ensures that the collected data captures different aspects of manipulation intelligence while maintaining high quality standards.

\subsection{Data Organization and Storage Format}

\begin{figure}[htbp]
\centering
\includegraphics[width=0.85\textwidth]{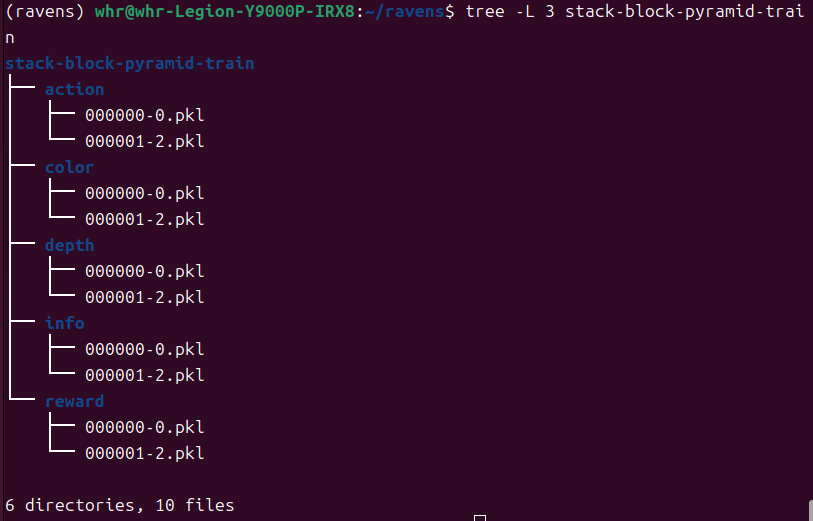}
\caption{PyBullet data organization structure with distributed storage format for efficient VLA training. 
Data modalities are separated into specialized directories (color, depth, action, reward, info) using \texttt{episode\_id-step\_id.pkl} naming convention, enabling efficient random access and supporting parallel processing for large-scale VLA dataset construction and training pipelines.}
\label{fig:data_structure}
\end{figure}

PyBullet data collection is based on Ravens' distributed storage structure, storing different modality data in independent directories. As shown in Figure~\ref{fig:data_structure}, this structure is organized by episodes for efficient access during training.

\begin{figure}[htbp]
\centering
\begin{subfigure}[t]{0.48\textwidth}
    \centering
    \includegraphics[width=\textwidth]{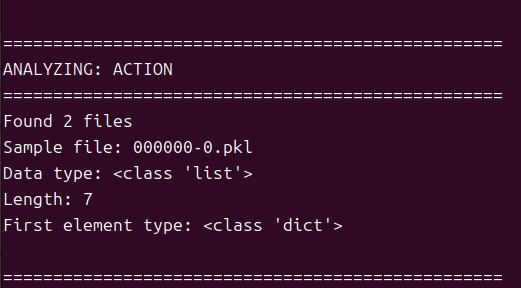}
    \caption{Action data structure}
\end{subfigure}
\hfill
\begin{subfigure}[t]{0.48\textwidth}
    \centering
    \includegraphics[width=\textwidth]{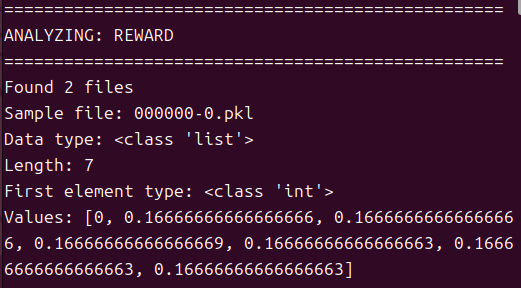}
    \caption{Reward computation results}
\end{subfigure}

\vspace{0.3cm}

\begin{subfigure}[t]{0.48\textwidth}
    \centering
    \includegraphics[width=\textwidth]{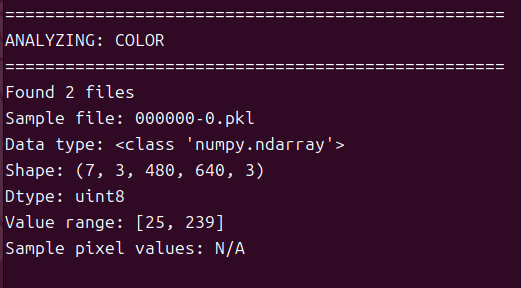}
    \caption{Color (RGB) data format}
\end{subfigure}
\hfill
\begin{subfigure}[t]{0.48\textwidth}
    \centering
    \includegraphics[width=\textwidth]{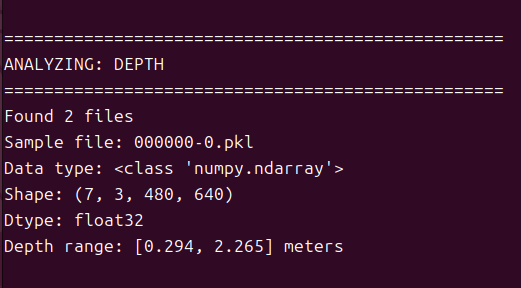}
    \caption{Depth sensor data}
\end{subfigure}

\vspace{0.3cm}

\begin{subfigure}[t]{0.6\textwidth}
    \centering
    \includegraphics[width=\textwidth]{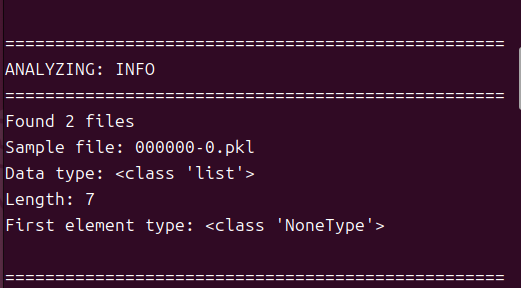}
    \caption{Info metadata structure}
\end{subfigure}

\caption{Actual data analysis results from stack-block-pyramid task showing real data structures, types, and values captured during PyBullet demonstration collection.}
\label{fig:data_analysis}
\end{figure}

To demonstrate the actual content and structure of each data modality, Figure~\ref{fig:data_analysis} shows the real data outputs from our stack-block-pyramid task collection.

The color directory stores RGB visual observations captured from PyBullet's camera system, as shown in Figure~\ref{fig:data_analysis}(c). Each file contains a (7, 3, 480, 640) numpy array with uint8 dtype, representing 7 timesteps of 480×640 RGB images. The value range [25, 239] indicates the pixel intensity distribution in the simulation environment. Action data, demonstrated in Figure~\ref{fig:data_analysis}(a), consists of lists containing 7 timesteps where each element is a dictionary with manipulation primitives and parameters computed by Ravens' oracle policies. The reward directory stores task completion signals following the progression pattern [0, 0.167, 0.167, 0.167, 0.167, 0.167, 0.167] as shown in Figure~\ref{fig:data_analysis}(b), where the initial timestep starts at 0 and subsequent timesteps receive incremental rewards of 0.167 based on block stacking progress.

Depth measurements are captured simultaneously with RGB images, containing (7, 3, 480, 640) float32 arrays that provide distance measurements ranging from 0.294 to 2.265 meters from the camera sensor to scene surfaces, as demonstrated in Figure~\ref{fig:data_analysis}(d). The info directory preserves comprehensive metadata including environment states, object poses, and task parameters, organized as lists of length 7 corresponding to the timestep sequence shown in Figure~\ref{fig:data_analysis}(e). Each file follows the naming convention \texttt{episode\_id-step\_id.pkl}, where the consistent 7-timestep structure across all modalities ensures temporal synchronization necessary for VLA training while enabling efficient random access during model development.

\section{Data collection with LIBERO}

\subsection{From Standard Benchmarks to Customized data}

LIBERO~\citep{qin2023libero} is a well-structured benchmark suite that defines standardized tasks and evaluation protocols using the robosuite~\citep{zhu2020robosuite} framework, which is built on the MuJoCo~\citep{todorov2012mujoco} physics simulator, providing modular environments specifically tailored for Vision-Language-Action (VLA) learning. Our work with LIBERO encompasses two main aspects: utilizing existing benchmark datasets and manually collecting new demonstrations through teleoperation. Each approach has its respective advantages and limitations.

First, we extensively leveraged LIBERO's existing demonstration datasets, particularly those curated in the cleaned versions released under the OpenVLA project~\citep{kim2024openvla}. These datasets remove no-operation actions, achieving denser and more meaningful state-action transitions. The cleaned datasets (\texttt{libero\_10\_no\_noops}, \texttt{libero\_goal\_no\_noops}, etc.) improve training stability by eliminating long stretches of idle steps that provide little to no learning signal, thereby ensuring that each time step contributes relevant supervision for policy learning.

However, relying solely on pre-collected data limits our ability to adapt tasks or enhance data diversity. Therefore, we also explored an alternative pathway: collecting customized trajectories using human teleoperation.

\begin{figure}[h]
\centering
\includegraphics[width=0.45\textwidth]{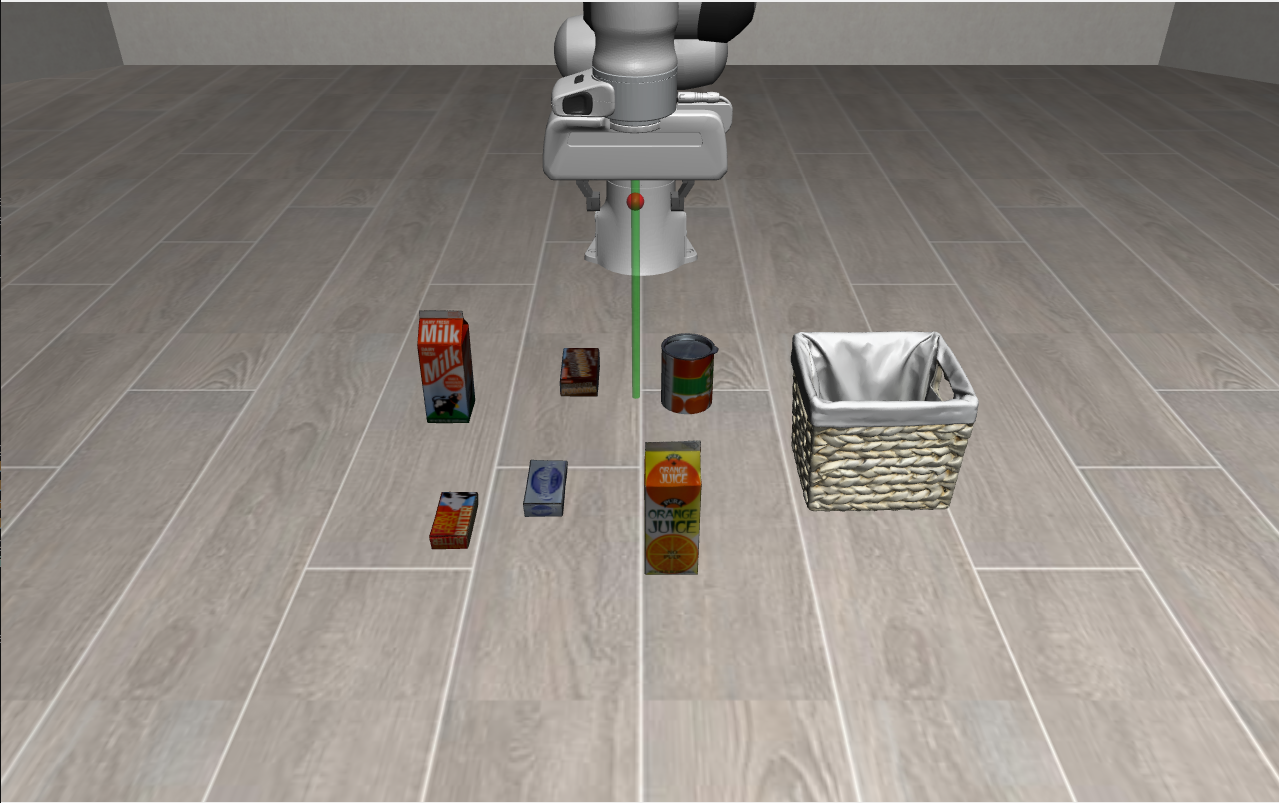}
\includegraphics[width=0.45\textwidth]{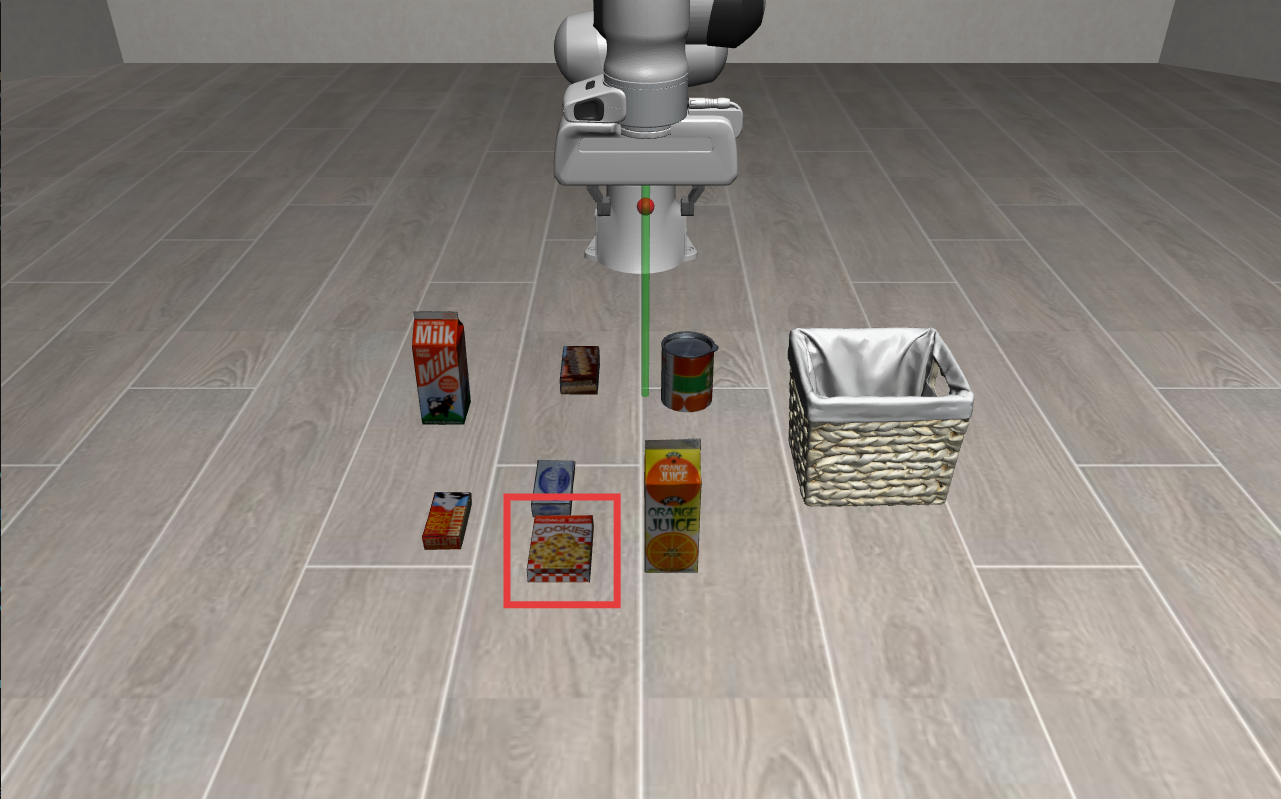}
\caption{LIBERO scene modification for new data collection experiments. 
(Left) Original standardized scene configuration for object manipulation tasks, (Right) Modified scene with additional distractor objects.}
\label{fig:scene_comparison}
\end{figure}

To evaluate robustness and create challenging scenarios, we modified LIBERO scenes by introducing additional objects that could potentially distract or mislead the robot during task execution. Figure~\ref{fig:scene_comparison} shows the comparison between original and modified scenes.

To implement these scene modifications, we developed custom BDDL (Behavioral Domain Definition Language) files to define new object locations, properties, and initialization logic. BDDL serves as LIBERO's standardized task specification format, enabling precise control over object placement, spatial relationships, and goal conditions within the simulation environment. Our approach primarily focused on exploring how to add pre-registered objects from LIBERO's built-in library, leveraging the existing object registry to create more complex and challenging scenarios.

\subsection{Built-in Object Integration}

Building on the scene modification approach shown in Figure~\ref{fig:scene_comparison}, we now detail the specific BDDL file modifications required to systematically add distractor objects. LIBERO's built-in object library contains pre-validated objects with established physics parameters, collision geometries, and visual properties, making them immediately suitable for the experimental complexity enhancement we demonstrated.

\begin{figure}[htbp]
\centering
\begin{subfigure}[t]{0.48\textwidth}
    \centering
    \includegraphics[width=\textwidth]{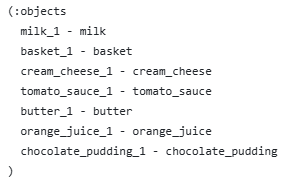}
    \caption{Original objects section}
    \label{fig:original_objects}
\end{subfigure}
\hfill
\begin{subfigure}[t]{0.48\textwidth}
    \centering
    \includegraphics[width=\textwidth]{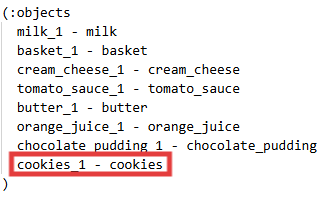}
    \caption{Modified objects section}
    \label{fig:modified_objects}
\end{subfigure}

\vspace{0.5cm}

\begin{subfigure}[t]{0.48\textwidth}
    \centering
    \includegraphics[width=\textwidth]{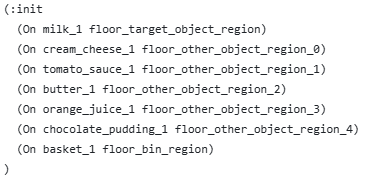}
    \caption{Original initialization}
    \label{fig:original_init}
\end{subfigure}
\hfill
\begin{subfigure}[t]{0.48\textwidth}
    \centering
    \includegraphics[width=\textwidth]{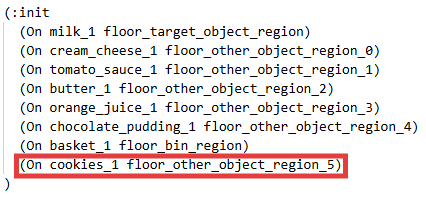}
    \caption{Modified initialization}
    \label{fig:modified_init}
\end{subfigure}

\vspace{0.3cm}

\begin{subfigure}[t]{\textwidth}
    \centering
    \includegraphics[width=0.6\textwidth]{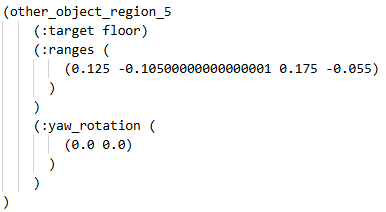}
    \caption{New spatial region definition (other\_object\_region\_5)}
    \label{fig:new_region}
\end{subfigure}

\caption{BDDL file modifications for adding distractor objects. (a-b) Objects section showing addition of cookies\_1. (c-d) Initialization section showing placement of new object. (e) New spatial region definition for distractor positioning.}
\label{fig:bddl_modification}
\end{figure}

Figure~\ref{fig:bddl_modification} shows the specific BDDL code changes that enable the scene transformation from Figure~\ref{fig:scene_comparison}. The modification process involves three coordinated steps that maintain the original task structure while introducing the additional visual complexity visible in the modified scene.

As shown in Figure~\ref{fig:bddl_modification}(a-b), we added \texttt{cookies\_1 - cookies} to the objects section to create the additional distractor visible in Figure~\ref{fig:scene_comparison}(Right). This leverages LIBERO's built-in object registry, which contains pre-validated cookies with established physics parameters and visual properties. No additional asset preparation or physics tuning is required, making this approach ideal for rapid experimental setup.

The initialization changes in Figure~\ref{fig:bddl_modification}(c-d) show the addition of \texttt{(On cookies\_1 floor\_other\_object\_region\_5)}, which places the distractor in the designated region visible in our modified scene. Crucially, \texttt{cookies\_1} remains excluded from the \texttt{(obj\_of\_interest )} specification, making it a passive element that doesn't trigger goal conditions or affect reward calculations. This design preserves the original task semantics while enabling the controlled complexity increase shown in Figure~\ref{fig:scene_comparison}.

Figure~\ref{fig:bddl_modification}(e) shows the new \texttt{other\_object\_region\_5} definition with coordinates \texttt{(0.125 -0.105 0.175 -0.055)} and \texttt{yaw\_rotation (0.0 0.0)}. This positioning strategy ensures that the distractor object enhances visual complexity without interfering with the core manipulation task, as demonstrated in the scene comparison. Careful placement is essential because LIBERO's fixed-position cameras generate 2D projective images without inherent depth separation. When objects appear close in the camera view, vision models may perceive them as connected components even if physically separate in 3D space, leading to incorrect target identification or manipulation attempts.

\subsection{Demonstration Collection}

\begin{figure}[htbp]
\centering
\includegraphics[width=\textwidth]{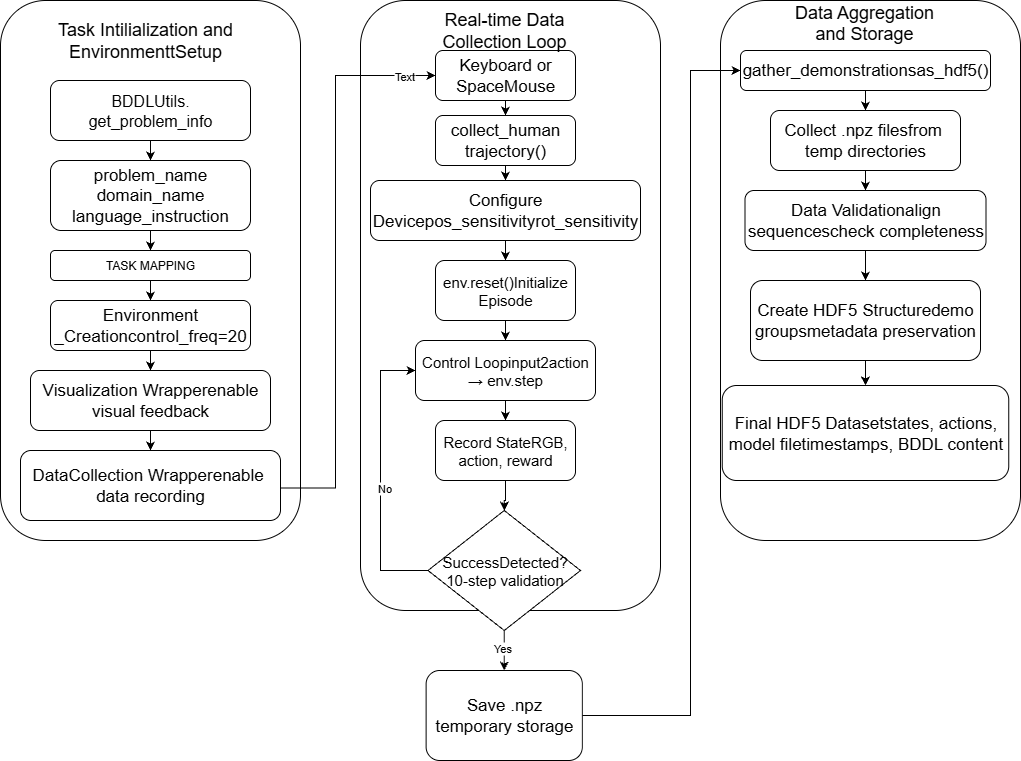}
\caption{LIBERO demonstration collection workflow showing the complete pipeline from BDDL task specification to HDF5 dataset generation. The process includes task parsing and environment setup (left), real-time human teleoperation with device input processing and success validation (middle), and data aggregation from temporary .npz files to structured HDF5 format (right). The success detection includes a 10-step validation mechanism to prevent false positives.}
\label{fig:libero_workflow}
\end{figure}

After completing scene configuration, we moved to the actual demonstration collection phase. Figure~\ref{fig:libero_workflow} illustrates our pipeline, which transforms abstract task descriptions into datasets through three interconnected stages.

\subsubsection{Task Initialization and Environment Setup}

The left panel shows how we convert human-readable task descriptions into executable robot simulations. We start with BDDL files—structured text files that describe robot tasks by specifying what objects exist, where they should be placed, and what constitutes successful task completion. The \texttt{BDDLUtils.get\_problem\_info()} function acts as our parser, reading these files and extracting three critical components: \texttt{problem\_name} identifies the specific manipulation task (such as ``put\_item\_in\_drawer''), \texttt{domain\_name} specifies the task category for system organization, and \texttt{language\_instruction} provides the natural language description that will be paired with robot actions for training.

\texttt{TASK\_MAPPING} serves as LIBERO's registry that automatically connects task names to their corresponding simulation environments. When you specify ``put\_item\_in\_drawer'', it knows to load a virtual world containing a robot arm, a drawer, appropriate objects, and the correct physics settings. This automation eliminates manual scene construction for each task type.

Environment creation includes setting \texttt{control\_freq=20}, which determines that the simulation updates 20 times per second. This frequency ensures smooth robot movements that feel natural to human operators—too low would create jerky motion, while too high would be computationally wasteful. Two essential wrappers are then applied: \texttt{VisualizationWrapper} provides the graphical interface that human operators see, while \texttt{DataCollectionWrapper} automatically captures and stores all demonstration data without requiring manual saving commands.

\subsubsection{Real-time Data Collection Loop}

The middle panel illustrates the core demonstration recording process. Human operators choose between keyboard or SpaceMouse input devices. Keyboard control creates discrete, step-like movements that produce jerky robot trajectories unsuitable for learning smooth manipulation. SpaceMouse is a specialized 3D device that translates physical force and torque into smooth, continuous robot movements across all six degrees of freedom, making it our preferred choice for high-quality demonstrations.

The \texttt{collect\_human\_trajectory()} function orchestrates the entire demonstration session, coordinating between human input, robot simulation, and data recording. Device configuration involves setting sensitivity parameters: \texttt{pos\_sensitivity} determines how much the robot moves in response to translational input, while \texttt{rot\_sensitivity} controls rotational responsiveness. Proper tuning ensures comfortable control without oversensitive or sluggish response.

Each demonstration episode begins with \texttt{env.reset()}, which initializes a fresh task scenario by resetting the robot to its starting position and randomizing object placements within appropriate ranges. This randomization ensures dataset diversity rather than the identical scenarios of multiple trials.

The real-time control loop operates through several synchronized steps. Human device input is processed by \texttt{input2action()}, which converts raw device signals into standardized robot control commands while handling coordinate transformations and safety constraints. These commands are executed through \texttt{env.step()}, which advances the simulation by one timestep, updating robot positions and calculating physics interactions. Simultaneously, the system records RGB camera images, executed actions, and reward signals, providing comprehensive multimodal data.

Success detection employs \texttt{env.\_check\_success()} to evaluate whether the current configuration satisfies task completion criteria defined in the BDDL file. However, to prevent false positives from brief object contact or accidental alignment, we implement 10-step validation: tasks are only considered complete when success conditions remain satisfied for 10 consecutive timesteps (0.5 seconds at 20Hz). This temporal requirement ensures genuine task completion. Successful demonstrations are saved as \texttt{.npz} files—a compressed format for efficiently storing numerical arrays.

\subsubsection{Data Aggregation and Storage}

The right panel shows how individual demonstration files become organized training datasets. The \texttt{gather\_demonstrations\_as\_hdf5()} function processes all \texttt{.npz} files from temporary directories, combining them into a single structured database suitable for machine learning training.

This aggregation process includes data validation to ensure trajectory completeness—checking that image sequences align with action sequences and filtering out demonstrations that end prematurely or contain recording errors. The HDF5 format organizes information hierarchically, with each demonstration becoming a separate group containing its complete sequence of states, actions, and model specifications. This structure enables efficient data access during training, allowing AI algorithms to quickly locate specific demonstrations.

Comprehensive metadata preservation includes collection timestamps, environment configuration parameters, and complete BDDL task descriptions. This metadata enables researchers to analyze how collection conditions affect data quality and ensures experiment reproducibility by allowing recreation of identical conditions using the preserved parameters.

\section{RT-X Dataset: Cross-Embodiment Learning for VLA}

\subsection{The Cross-Embodiment Learning Paradigm}

The RT-X dataset represents a fundamental shift from single-robot learning to cross-embodiment learning. Traditional robotic datasets typically focus on individual robot platforms, thereby limiting their applicability across different hardware configurations. RT-X addresses this limitation by aggregating data from 22 different robot embodiments across 21 institutions~\citep{open2023rt}, creating a unified dataset of over 1 million real robot trajectories. The key insight is that with appropriate abstraction, manipulation skills can be effectively transferred across different robot modalities, enabling a single model to control multiple robot types.

\subsection{Dataset Composition and Representative Examples}

\begin{figure}[htbp]
\centering
\includegraphics[width=\textwidth]{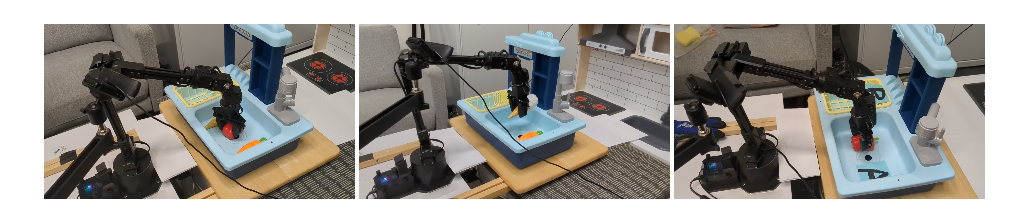}
\caption{Representative manipulation tasks from the RT-X dataset~\citep{open2023rt}. Top row shows typical manipulation scenarios: moving objects to specific locations (``move red pepper to tray''), grasping operations (``pick ice cream''), and precise placement tasks (``move red pepper to A''). Figure adapted from Open X-Embodiment.}
\label{fig:rtx_examples}
\end{figure}

RT-X encompasses diverse manipulation scenarios across different robot platforms, environments, and tasks~\citep{open2023rt}. Figure~\ref{fig:rtx_examples} shows representative data samples from the dataset, illustrating the variety of robots, scenes, and manipulation skills captured across real-world settings.

The dataset composition has important practical implications for VLA training. The Franka Panda dominates with 40\% of all trajectories, reflecting its widespread adoption in research institutions due to its versatility and research-friendly interface. This concentration provides sufficient data for robust model training on this popular platform. WidowX robots contribute 25\% of the data, offering an important counterpoint with different kinematic structures and workspace constraints. Specialized platforms including xArm, Google Robot systems, and Sawyer robots provide the remaining 35\%, ensuring diversity necessary for effective cross-embodiment learning.

From a task perspective, pick-and-place operations constitute 60\% of all demonstrations, making RT-X particularly well-suited for basic manipulation research. The remaining 40\% includes more complex behaviors: wiping motions, assembly tasks, tool use, and container manipulation. While these advanced skills have smaller sample sizes, they provide essential coverage for developing more sophisticated manipulation capabilities. The dataset spans over 300 distinct scenes ranging from controlled laboratory environments to realistic kitchen settings, with comprehensive object diversity covering geometric shapes, containers, food items, and household appliances.

\subsection{Data Format and Practical Usage}

RT-X employs a crucial technical innovation: standardized action representation. All robot actions are stored in a unified 7-dimensional format [x, y, z, roll, pitch, yaw, gripper], representing end-effector movement and gripper state. This abstraction means that regardless of whether the original data came from a 6-joint WidowX or 7-joint Franka robot, all actions are converted to this common format that captures the essential manipulation semantics.

This standardization eliminates a major barrier in multi-robot learning and provides RT-X's primary advantage: unprecedented combination of scale and diversity within a consistent format. When using RT-X data, researchers receive these normalized actions along with corresponding RGB images and language instructions, enabling direct training without robot-specific preprocessing. Rather than investing months in data collection infrastructure, researchers can immediately access million-scale trajectories covering common manipulation scenarios. This accessibility is particularly valuable given that robot data collection is orders of magnitude more expensive and time-consuming than traditional curation of vision and language datasets.

RT-X demonstrates different strengths for different research objectives. For basic manipulation research, particularly pick-and-place capabilities, the 60\% task coverage provides substantial training data enabling vision language action model development. For specialized applications like assembly or tool use, RT-X serves as a foundation that researchers can extend with task-specific data collection. The multi-institutional nature introduces quality variations across collection sites, but this variation can actually improve model robustness by providing exposure the suboptimal demonstrations that models will encounter in real-world deployment.

\section{Conclusion}

Our analysis reveals that effective VLA dataset construction requires careful consideration of three key dimensions: data quality, scale, and diversity. Simulation-based approaches like PyBullet excel at providing controlled, high-quality demonstrations with perfect annotations, making them ideal for algorithmic development and systematic studies. Human-guided methods like LIBERO offer realistic demonstrations that better reflect real-world conditions, though with higher collection costs and variable quality. Large-scale collaborative efforts like RT-X achieve unprecedented diversity and real-world applicability through multi-institutional data collaboration, albeit with inherent quality variations across collection sites.

Several technical considerations are crucial for successful data collection. Automated data generation requires robust oracle policies and systematic quality control, while human teleoperation demands careful validation mechanisms such as multi-step success verification to ensure demonstration authenticity. Cross-platform compatibility depends on standardized data representations that abstract away hardware differences, enabling knowledge transfer across different robot embodiments.

\bibliography{iclr2024_conference}
\bibliographystyle{iclr2024_conference}

\end{document}

%% file: math_commands.tex
\usepackage{amsmath,amsfonts,bm}

\def\eqref#1{equation~\ref{#1}}

\def\1{\bm{1}}

\DeclareMathAlphabet{\mathsfit}{\encodingdefault}{\sfdefault}{m}{sl}
\SetMathAlphabet{\mathsfit}{bold}{\encodingdefault}{\sfdefault}{bx}{n}

